\documentclass[conference]{IEEEtran}
\ifCLASSINFOpdf
\else
\fi

\usepackage{soul}
\usepackage{graphicx,array}
\usepackage{cite}
\usepackage{amsmath,amssymb,amsfonts}
\usepackage{graphicx}
\usepackage{textcomp}
\usepackage{algorithm}
\usepackage{algpseudocode}
\usepackage{color,soul}
\usepackage{diagbox}

\usepackage{verbatim}
\usepackage[breaklinks=true]{hyperref}
\usepackage{cite}
\usepackage{textcomp}
\usepackage[table]{xcolor}
\usepackage{enumitem}
\usepackage{multicol}
\usepackage{multirow}
\usepackage{gensymb}
\usepackage{xcolor}
\usepackage[T1]{fontenc}
\usepackage[autostyle]{csquotes}

\newcommand\ivan[1]{\textcolor{orange}{Ivan : #1}}
\newcommand\subash[1]{\textcolor{cyan}{Subash : #1}}

\usepackage{algorithm}
\usepackage{algpseudocode}

\usepackage{bm}

\usepackage{booktabs,tabularx,enumitem,ragged2e}
\usepackage{colortbl}
\usepackage[utf8]{inputenc}

\begin{document}
% Privacy Attacks Against AI-Robots utilizing Digital Twins
% Securing AI and Digital Twins in Robotics: A Survey of Privacy Attacks
% Privacy Attacks Against Robots Enabled by AI and Digital Twins
%\title{Securing Digital Twin Systems in AI-Robotics: \\A Survey of Privacy Attacks}  %(Deadline: 14 April 2024 (rough draft), 8 page limit for IEEE CIC)
\title{A Survey on Privacy Attacks Against Digital Twin Systems in AI-Robotics}

% \title{Privacy Attack on AI-Enabled Robotics Digital Twin: A Growing Threat}

\author{\IEEEauthorblockN{Ivan A. Fernandez\IEEEauthorrefmark{1}, Subash Neupane\IEEEauthorrefmark{2},
Trisha Chakraborty\IEEEauthorrefmark{3},
Shaswata Mitra\IEEEauthorrefmark{4}, \\
Sudip Mittal\IEEEauthorrefmark{5},
Nisha Pillai\IEEEauthorrefmark{6},
Jingdao Chen\IEEEauthorrefmark{7},
Shahram Rahimi\IEEEauthorrefmark{8}}

\IEEEauthorblockA{Computer Science and Engineering,
Mississippi State University\\
Email: \{\IEEEauthorrefmark{1}iaf28,
\IEEEauthorrefmark{2}sn922,
\IEEEauthorrefmark{3}tc2006,
\IEEEauthorrefmark{4}sm3843\}@msstate.edu,
\{\IEEEauthorrefmark{5}mittal, \IEEEauthorrefmark{6}pillai, \IEEEauthorrefmark{7}chenjingdao,
\IEEEauthorrefmark{8}rahimi\}@cse.msstate.edu}
}


%Telephone: (800) 555--1212, Fax: (888) 555--1212}


\maketitle


% \hl{none of the acronyms have full forms}

\begin{abstract}
Industry 4.0 has witnessed the rise of complex robots fueled by the integration of Artificial Intelligence/Machine Learning (AI/ML) and Digital Twin (DT) technologies. While these technologies offer numerous benefits, they also introduce potential privacy and security risks. This paper surveys privacy attacks targeting robots enabled by AI and DT models. Exfiltration and data leakage of ML models are discussed in addition to the potential extraction of models derived from first-principles (e.g., physics-based). We also discuss design considerations with DT-integrated robotics touching on the impact of ML model training, responsible AI and DT safeguards, data governance and ethical considerations on the effectiveness of these attacks. We advocate for a trusted autonomy approach, emphasizing the need to combine robotics, AI, and DT technologies with robust ethical frameworks and trustworthiness principles for secure and reliable AI robotic systems.
\begin{comment}
\hl{Industry 4.0 has witnessed the rise of Digital Twin (DT)-integrated  robots in safety-critical applications, fueled by the integration of Artificial Intelligence (AI) and DT technologies. While these technologies offer numerous benefits, they also introduce potential privacy and security risks. This paper surveys privacy attacks from security standpoint targeting DT-integrated intelligent robots, focusing on data exfiltration via model inference APIs, cyber means and LLMs, including vulnerabilities in both ML-enabled and Physics-based models. We also discuss the impact of ML model training, responsible AI safeguards, and ethical considerations on the effectiveness of these attacks. We advocate for a trusted autonomy approach, emphasizing the need to combine robotics, AI/ML, and DT technologies with robust ethical frameworks and trustworthiness principles for secure and reliable intelligent robotic systems.}
\end{comment}

\end{abstract}


% \ivan{IEEE CIC 2024, Washington, DC \\ https://www.sis.pitt.edu/lersais/conference/cic/2024/calls.html \\
% Research track should be 10 pages \\
% Submission deadline: May 30}

% \fi
\IEEEpeerreviewmaketitle

\begin{figure*}[h]
\centering
\includegraphics[scale=.65]{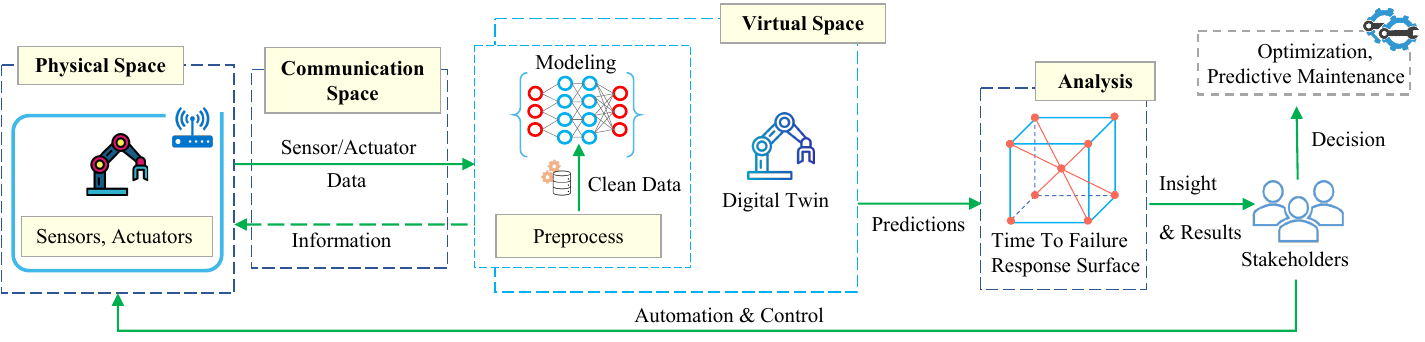}
\caption{Generic Framework of Data-Driven Robotic Digital Twin: Physical spaces comprise robotic sensors that collect data. Virtual space utilizes the data collected from physical space via a communication link between them. Predictions are generated by the AI models within vitual space, which are then analyzed before decisions are made by stakeholders.}
%\ivan{Subash can you add some context to the caption?}
    \label{fig: DT_architecture}
    \vspace{-3mm}
\end{figure*}

\section{Introduction and Motivation}

\begin{comment}
% From Dr. Mittal
IR4
AI+Robots
Critical apps

Security Issues + Example
Privacy Attacks + Example

This paper.. 

Contributions 
\end{comment}

% \hl{Remove focus from CPS. DT is not defined, the para should be motivating Robots +DT}

% \hl{the opening is not scaring me enough. A good idea for this kind of paper is to scare the reader. how about a scary scenario about compromized DT security? what impact will it have on bussiness operations and trust? DTs are used in what industries? } 

% \hl{imagine the DTs of space satellites getting compromised. OR imagine the DTs in aerospace industry getting compromised. }

%With the emergence 
% In the modern era of Industry 4.0, the rise of intelligent robots combined with Digital Twin(DT) technologies have revolutionized  various industries, including energy, transportation \cite{neupane2023twinexplainer}, manufacturing, and healthcare among other domain. However, this technological advancement bring forth significant cybersecurity challenges that threaten the integrity and reliability of these systems. 
% al sectors have seen an increase in automation and self-monitoring by fusing technologies such as  Artificial ntelligence (AI), robotics, cloud computing, big data analytics, Internet-of-Things (IoT), and cyber-physical systems (CPS). 

% \hl{made slight modifications in first para...to tone it down..also brifely stated what DT's are..Subash}

In today's interconnected world, the potential impact of cyberattacks on critical infrastructure cannot be overstated. Take, for example, a manufacturing facility brought to a standstill, not by a physical breakdown, but by a carefully orchestrated cyberattack. The target: the Digital Twin (DT) (virtual representations of physical systems)\cite{neupane2023twinexplainer} of a critical robotic arm, which is an automated machine used for a complex assembly process. An adversary, exploiting vulnerabilities in the DT's Artificial Intelligence (AI) system, manipulates sensor readings, feeding false information about the robot's position and environment. The consequences of such a breach extend beyond operational disruptions (for example, production grinds to a halt, costly repairs, maintenance etc.); they undermine business operations and trust, posing serious risks to safety-critical environments where human lives might be at stake. This scenario, unfortunately, is not science fiction. 
With the increasing reliance on AI-enabled robotic DT systems across industries such as military\cite{metcalf2023integrating}, aerospace\cite{xiong2021digital}, intelligent manufacturing\cite{xu2019digital}, healthcare \cite{croatti2020integration}, smart cities and transportation \cite{el2020roads, neupane2023twinexplainer}, the security of their digital twins becomes paramount.

%Industry 4.0 is known as \emph{The Age of Cyber-Physical Systems} because physical systems have now become more closely integrated with digital technologies. By expanding on information and communication technologies (ICT) \cite{xu2018industry}, CPS is the integration of computing devices, actuation and control, networking infrastructure, and sensing of the physical world. The National Institute of Standards and Technology (NIST) \cite{greer2019cyber} defines CPS as ``engineered systems that are built from, and depend upon, the seamless integration of computation and physical components.''  There are many domains making use of recent advances in CPS including smart homes, smart cars, smart grids, medical devices, and supervisory control and data acquisition (SCADA) systems \cite{yaacoub2020cyber}. In this paper, we investigate the cybersecurity risks of intelligent robots, or complex CPS, that make use of AI and digital twin (DT) technologies. 

The convergence of Industry 4.0 technologies - AI, robotics, cloud computing, and the Internet-of-Things (IoT) - has given rise to sophisticated Cyber-Physical Systems (CPS). At the heart of this revolution are DTs. % – virtual representations of physical assets that mirror their real-world counterparts in near real-time. 
By leveraging AI and DT technologies, AI robots are now capable of higher levels of autonomy and complex Human-Robot Interactions (HRI) within multi-agent environments \cite{neupane2023security}, finding applications in safety-critical domains where human lives, system integrity, and environmental safety are at stake \cite{mazumder2023towards}.
% Intelligent robots are used in many applications aiding humans in various industries including manufacturing, health care, and transportation. AI and DT methodologies have been key enablers for intelligent robots \cite{mazumder2023towards} allowing them function with some level of autonomy in safe-critical applications where human-life, the system, or the environment may be at risk.
% % \trisha{Are there other type?} \ivan{addressed.}.
% The next generation of intelligent robots will be capable of higher levels of autonomy and sophisticated human-robot interactions (HRI) within multi-agent environments \cite{neupane2023security} by making use of AI and DT technologies. 
% \hl{why is the security of DTs a unique problem? What bad thing will happen if this security is compromized?}
However, this reliance on AI and DTs introduces significant security risks, particularly in the realm of privacy. AI-enabled robotic DT models inherently rely on vast amounts of data, making them susceptible to breaches if inadequate data governance and cybersecurity measures are in place. An unsecured communication protocol, for instance, can provide unauthorized access to sensitive data and algorithms ~\cite{botta2023cyber,cerrudo2017hacking}, potentially compromising the entire system. The case of the Aethon TUG, a smart autonomous mobile robot used in hospitals, highlights this vulnerability. Researchers discovered critical vulnerabilities in the system that could allow adversaries to seize control and extract sensitive patient information \cite{Cynerio}.

This vulnerability is further amplified by the tight coupling between AI and DTs. Robotic DT models often incorporate AI models for analysis and decision-making, creating a complex ecosystem where a security flaw in one component can cascade into the other. Recognizing these risks, governments and regulatory bodies are developing guidelines for Responsible AI (RAI). The United States White House, for instance, issued an executive order in October 2023, establishing new standards for AI safety and security \cite{biden2023executive}.

While the security risks of AI and DTs have been studied independently, the synergistic impact of their convergence on AI robotic systems, particularly from a privacy perspective, remains largely unexplored. This paper addresses this critical gap, investigating the unique challenges and vulnerabilities arising from the integration of AI and DTs in AI robots.
To better understand and mitigate these emerging threats, we will utilize the MITRE Adversarial Threat Landscape for Artificial Intelligence Systems (ATLAS) \cite{atlasadversarial} framework. While ATLAS provides a valuable resource for understanding AI-specific threats, we argue that it needs to be extended to adequately address the privacy risks associated with DTs. The main contributions of this paper are as follows:

   \begin{enumerate}
   
    \item  Provide a survey of privacy attacks against DT systems used by AI robots.
    \item Suggest the addition of first-principles models to the MITRE ATLAS framework under exfiltration.
    % \hl{in a previous para you need to introduce ATLAS, then argue its not enough for DTs}
    \item Provide a discussion on how trusted autonomy can be achieved by combining robotics, AI, and DT technologies with ethics and trustworthiness concepts. 

   \end{enumerate}

The remainder of the paper investigates the security of AI and DT systems in robotics. In Section \ref{background} we provide a primer on digital twin and then discuss its integration with robotics. Section \ref{attacks} explores the attack surfaces on robotics systems that make of use AI and DT models. Following that, we touch on the impact of using machine learning (ML) models, responsible AI and DT safeguards, data governance and ethical consideration for DT-integrated robotics in Section \ref{discussion}. Finally, Section \ref{conclusion} concludes our paper.

\section{Background}
\label{background}

In this section we provide the background on Digital Twin (DT) technology, and discuss the 
% Intelligent robots can leverage DT models to safely operate in complex environments. In this section, we discuss the
impact of digital twins to robotics including the importance effective data exchange.

\subsection{Digital Twin Paradigm}

% \ivan{Subash, please address comments by Trisha in this section.}
% \subash{Addressed}

The concept of using \enquote{physical twins}, which served as an early precursor to digital twins, has its historical origins in the 1970s, aligning with NASA's Apollo mission \cite{neupane2023twinexplainer}. In 2002, Grieves et al. \cite{grieves2014digital} introduced the notion of DTs informally, later formalizing it in their published white paper. In 2012, NASA and United States Air Force (USAF) researchers \cite{glaessgen2012digital} define a digital twin as \textit{``an integrated multiphysics, multiscale, probabilistic simulation of
an as-built vehicle or system that uses the best available physical models, sensor updates, fleet history, etc., to mirror the life of its corresponding flying twin.''}  
% \trisha{Nitpicking: use similar type of quotations ("" or ``'')} \ivan{Addressed.}
A simpler definition is that a DT is a  virtual prototype of physical assets that simulates, emulates, mirrors, or twins the real-time operational conditions to behave like a real physical asset. {A typical DT model as depicted in Fig.} \ref{fig: DT_architecture} {comprises three components: a \textit{physical space, virtual space,} and \textit{communication space}. We explain the functions of each these components in greater details below.}  %\trisha{Refer to Figure 1, looks like you are explaining each component of Figure 1 here.}

\subsubsection{Physical Space}
%The physical space is comprised of real-world objects, such as things, equipment, systems, \hl{cameras, sensors}, components etc. that are responsible for collecting data of current physical measurement of objects. 
{The physical space is comprised of real-world objects, such as equipment, systems, cameras, sensors, components, etc. that are responsible for collecting data of the current physical measurement of objects.}
In the context of robotics systems, the two fundamentally connected physical components are sensors and actuators, which often work in tandem but are essentially opposite in their nature. For example, a sensor monitors the state and sends a signal when changes occur, whereas an actuator receives the signal and performs an action. Robots today are outfitted with several sensors that generate volumes of operational data. These data are typically collected in time-series format, expressed as multi-channel sensor data, and stored in a datastore \cite{neupane2022temporal}.

\begin{comment}
The data is then used to develop the DT-based system, through the application of AI models, statistical and probabilistic techniques, or mathematical models.
\end{comment}

\subsubsection{Virtual Space}
{In a DT system,  virtual space can be viewed as a virtual replica that maintains a real-time model of its physical space.}
%The data collected in the physical space is the input to this phase, where the data is preprocessed. The preprocessing step includes various sub-steps such as cleaning, removing redundant observations, transformation, data restructuring, and scaling. The output of the preprocessing step is a high-quality dataset. 
{It receives the data collected in the physical space as input which is then preprocessed. The preprocessing step includes various sub-steps such as cleaning, removing redundant observations, transformation, data restructuring, and scaling. The output of the preprocessing step is a high-quality dataset.}  Deep learning models and/or first-principles models are then applied to this data in order to create the DT-based system. %DT systems crafted in this manner have various capabilities, such as time-series forecasting \trisha{provide some citations} on the remaining useful life of a degradable vehicle component, continuous monitoring, abnormal patterns, and fault detection in data signals, etc.
DT systems crafted in this manner have various capabilities. They can visualize the instant status of their physical space \cite{luan2021paradigm} through continuous monitoring, perform time-series forecasting on the remaining useful life \cite{feng2023digital} of a degradable vehicle component, detect abnormal patterns \cite{neupane2023twinexplainer}, and perform fault detection in data signals. %\subash{reworked and added necessary citations}

\subsubsection{Communication Space}
The communication space is the infrastructure that connects physical and virtual spaces \cite{botin2022digital}. It allows for information exchange between the various components of the overall DT ecosystem. Digital twins require effective communication to maintain synchronization and accurately interact with their real-world counterparts. Both wired (e.g., fiber optics, CAN Bus, ARINC-429) and wireless (e.g., WiFi, Zigbee, Bluetooth, 5G) may be used for data exchange.

% \trisha{Figure 1 needs to be upgraded to incorporate communication space entity} \ivan{Subash is updating.} \subash{updated}

\subsubsection{Analysis}
%\ivan{Insights and Results. Tied back to Fig 1.}
Once the digital twin is produced, the insights generated by it can be used to achieve the given robotic systems goals, such as repairs, prognostics, optimization, or predictive maintenance. In addition, the results can be utilized to update and improve both the physical and virtual twins, which will provide more autonomy and control.

% \trisha{Do we want to follow Figure 1? In that case, need another subsection of how the insights and results are utilized}

\subsection{Digital Twin-Integrated Robotics}

An AI robot is an agent that can sense and act to maximize its chances of success for a given task \cite{murphy2019introduction}. Robots can have varying levels of autonomy (e.g., no autonomy, semi-autonomy, full-autonomy) based on how much human control is needed \cite{melenbrink2020site}. A fully autonomous AI robot is realizable by incorporating AI \cite{ness2023synergy}.  Furthermore, for an autonomous robot to be considered AI-enabled, it must demonstrate at least one of the fundamental constructs of AI such as \emph{reasoning, planning, learning, communication,} and \emph{perception} \cite{samoili2020ai}. 

\begin{figure}[htbp]
\centering
\includegraphics[scale=.71]{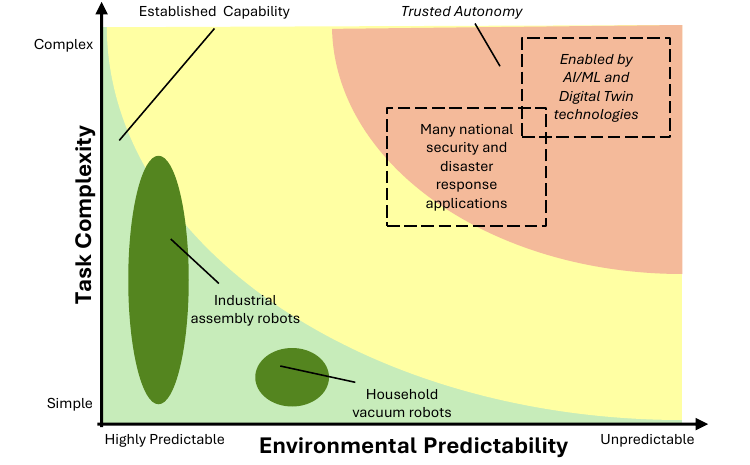}
\caption{Overview of autonomous capabilities adapted from Boulet et al. \cite{boulet2017autonomous}. Two dimensions of the Autonomy Levels for Unmanned Systems (ALFUS) model \cite{huang2005framework} are shown. We make two key additions to the original: (1)   \enquote{Future Capability} is replaced with \enquote{Trusted Autonomy} and (2) note that AI robots performing complex tasks in unpredictable environments must be enabled by some combination of AI/ML and DT technologies. }
    \label{fig: ALFUS}
\end{figure} 

The concept of a DT is complementary methodology to AI \cite{fuller2020digital, rathore2021role}. We argue that both AI and DT technologies are needed to achieve \emph{Trusted Autonomy} as shown in Figure \ref{fig: ALFUS}. AI aids with reducing the dimensionality of highly complex tasks and digital twin concepts facilitate exploration of different scenarios (or \emph{What-If simulations} \cite{pires2021digital}) safely and efficiently.

DTs offer a powerful tool for designing, testing, operating, and maintaining robotic systems more effectively. They can be integrated with Model-Based Design (MBD) \cite{bachelor2019model} and Model-Based Systems Engineering (MBSE) \cite{madni2019leveraging} processes to improve decision making throughout the robotic system's lifecycle. Lie et al. \cite{liu2021review} provide industrial applications for DTs in the design, development (manufacturing), service, and retirement phases of the lifecycle. During the design phase, DTs can be used for iterative optimization of the product design and support of virtual prototyping. Similarly, during the the manufacturing phase, the use of digital twins allows for real time monitoring and optimization of processes. Furthermore, DTs can be used for state monitoring, fault detection, predictive maintenance, and virtual testing during the service phase of the system.

DTs can be used to represent a robotic component, a robotic system, or a System of Systems (SoS). DTs can also be employed at the edge \cite{girletti2020intelligent} for real-time decision making or in fog/cloud infrastructures if the use case allows for some latency. The rest of this section provides examples of DTs from different tiers with respect to the robotic system.

\subsubsection{Component-level DTs}
The DT paradigm allows for modeling of specific components in complex systems \cite{zheng2022emergence} like AI-enabled robots. Component-level DTs can be used as virtual models in lieu of physical hardware to test out requirements. Kutzke et al. \cite{kutzke2021subsystem} discuss how digital twins can be used for subsystems in Autonomous Underwater Vehicles (AUV) by establishing a generic process for determining a set of priority-based system components requiring digital twin development for Condition-Based Maintenance (CBM) purposes.  In order to create a testbed to optimize electric propulsion drive systems in autonomous vehicles, Rassolkin et at. \cite{rassolkin2019digital} develop DTs using physical models and virtual sensors.

\subsubsection{System-level DTs}
Component-level DTs can be integrated and combined to model entire complex systems \cite{jia2023simple}. As an example, Xiong et al. \cite{xiong2022design} design and implement a simulation method for car-following scenarios of an autonomous vehicle using a digital twin for the secondary vehicle.

\subsubsection{SoS-level DTs}
Similar to developing a system DT with virtual models for its components, a System-of-Systems (SoS) DT can be constructed by aggregating multiple system-level models. DT modeling can be simple or highly complex, and Autonomous Mobile Robots (AMR) are usually modeled as complex SoS DTs \cite{stkaczek2021digital}.

\begin{comment}
\input{table_dt_intro}
\end{comment}

Adept adversaries can target any and all of the DT tiers. It is important to note that while information may be isolated at a component-level, if that component is vital to operation of the system, then a successful exfiltration by an adversary compromises the entire the system. Similarly, if a system within a SoS becomes victim of an attack, then the entire SoS is potentially compromised as well. 

\begin{comment}
In the next section, we discuss privacy attacks against intelligent robots that make use of digital twin methodologies.
\end{comment}

\section{Attacks on DT-Integrated AI Robots}
\label{attacks}

Digital twins provide an enabling technology for developing and safeguarding advanced robotic systems. As discussed in the previous section, digital twins allow for exploration of \enquote{What-If?} scenarios and analyses that may be prohibitive with the physical system since they can simulate the system's behavior under different conditions (e.g., environmental, system configuration). The DT is comprised of software components (e.g., algorithms and models) \cite{alcaraz2022digital} and unauthorized access to these digital assets could lead to devastating damage to stakeholders including the theft of Intellectual Property (IP), Private Personal Information (PPI), and reverse engineering.

% \begin{figure*}[htbp]
% \centering
% \includegraphics[scale=.7]{figs/attack.pdf}
% \caption{XXXXX\hl{First cut ready for review---Subash} }
%     \label{fig: attacks}
% \end{figure*} 

\begin{figure}[htbp]
\centering
\includegraphics[scale=.78]{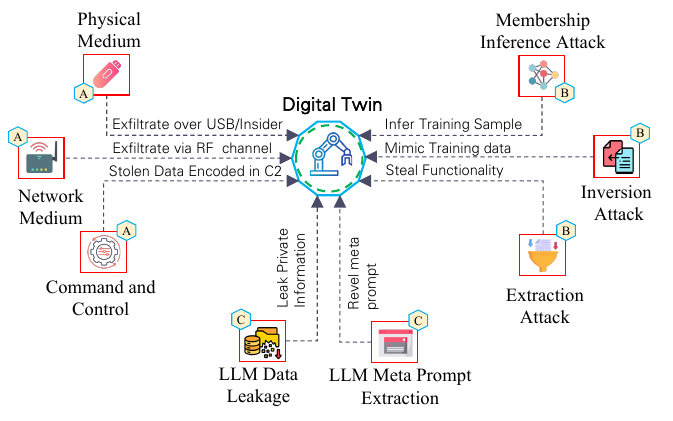}
\caption{A graphical illustration of different privacy exfiltration techniques. \textit{A} represents \textit{exfiltration via cyber means}, \textit{B} represents \textit{exfiltration via Model inference API}, while \textit{C} represents possible attacks within LLM space.}
    \label{fig: attacks}
\end{figure}

\begin{comment}
\begin{figure*}[htbp]
\centering
\includegraphics[scale=.65]{figs/attack.png}
\caption{A graphical illustration of different privacy exfiltration techniques. \textit{A} represents \textit{exfiltration via cyber means}, \textit{B} represents \textit{exfiltration via Model inference API}, while \textit{C} represents possible attacks within LLM space.}
    \label{fig: attacks}
\end{figure*}
\end{comment}

The attack surface of a robotic system increases if DT or AI models are used. Adversaries could launch cyberattacks if the proper cybersecurity mechanisms (e.g., secured IoT networks and protocols) are not in place for protecting these systems. The MITRE Adversarial Threat Landscape for Artificial-Intelligence Systems (ATLAS) \cite{atlasadversarial} provides a knowledge base of tactics, techniques, and procedures (TTPs) against AI-enabled systems. There are 14 major tactics in the ATLAS but the main focus for this paper lies with \textit{exfiltration}.

% The use of AI only increases the attack surface of DT-integrated intelligent robots. A DT without AI would be vulnerable to traditional attacks included in the MITRE Adversarial Tactics, Techniques, and Common Knowledge (ATT\&CK) \cite{strom2018mitre} framework.

%The 14 major tactics in ATLAS are \emph{reconnaissance, resource development, initial access, ML model access, execution, persistence, privilege escalation, defense evasion, credential access, discovery, collection, ML attack staging, exfiltration,} and \emph{impact}. This paper focuses on exfiltration.

Malicious actors use privacy attacks to steal information about a physical system if they are able gain access to its DT. If the DT is comprised of a ML model, then an adversary could attempt to extract the model data (e.g., architecture, weights) or the training data (e.g., confidential data). This includes large-language models (LLMs) \cite{evertz2024whispers} if they are used by the AI robot. Privacy attacks can also occur against physics-based (first-principles) models. An adept adversary could query a physics-based model to generate enough input/target pairs to a train a surrogate data-driven model. Privacy attacks are also known as confidentiality attacks because of their focus in the Confidentiality-Integrity-Availability (CIA) security triad.

Privacy attacks against DT-integrated AI robots could lead to catastrophic outcomes in safety-critical applications. This is because unauthorized access to a DT could lead to vulnerabilities in the physical twin (system) being leaked. Furthermore, allowing unauthorized access to sensitive information contained in a AI robot's digital twin could lead to additional attacks (cyber or physical) and extortion.

Robot vacuums fall under the simple task complexity and highly predictable region in Figure \ref{fig: ALFUS}. Their relatively known components and mass consumption can make them prime candidates for privacy attacks by adversaries. There are examples of these robots being used for eavesdropping via their LiDAR sensors \cite{zubatiuk2021development}. In scenarios where AI robots must operate in unpredictable environments and perform a highly complex tasks, privacy attacks can lead to catastrophic damage. With the adoption of IoT concepts in the military domain   \cite{pradhan2020security} (i.e., Internet-of-Military-Things (IoMT)), autonomous military vehicles are also vulnerable to attacks.  This was seen in December 2011, when Iranian hackers were able to bring down a RQ-170 Sentinel US spy drone \cite{yahuza2021internet}.

\begin{comment}
scenario, military assets could be defeated by an adversary capable of exfiltrating sensitive data and models contained in digital twins.
\end{comment}

The rest of this section will focus on privacy attacks against DT-integrated AI robots. In particular, we will discuss four types: \textit{Exfiltration via Model Inference API, Exfiltration via Cyber Means, LLM Meta Prompt Extraction}, and \textit{LLM Data Leakage} as depicted in Figure \ref{attacks}.  Table \ref{table:dt_attacks} provides a summary of these attacks.

\subsection{Exfiltration via Model Inference API}

Exfiltration is a form of data theft that occurs when an adversary steals artifacts and information about the AI robot from the DT ecosystem. Even though the use of Application Programming Interfaces (APIs) can to hide details of the model, exfiltration via model inference is still possible through black-box query access. In fact, this technique can be further divided into the following categories: \emph{training data membership inference, model inversion,} and \emph{model extraction}.

In Membership Inference Attacks (MIAs), adversaries aim to determine if a training sample belongs to the training data of the targeted ML model. MIAs can target various ML models (e.g., classification, generative models \cite{zhang2020secret}) and are usually constructed using one of two major approaches: binary classifier-based attack and metric-based attack \cite{hu2022membership}. In a binary classifier-based attack, \emph{shadow training} \cite{shokri2017membership} is usually used to train several shadow models as surrogate to the target model. Then an attack model (binary classifier) is trained to infer if a particular sample belonged to the target training data (member vs. non-member) from information provided by shadow models. Given a sample, metric-based inference attacks compares the output probability vector (confidence scores) provided by model against a predefined threshold as a way to determine if it belonged in the training data.

Expanding on metric-based inference attacks leads to label-only model inversion \cite{zhu2022label} where the goal is to infer sensitive information about the training data (or individual points) from output labels provided by the ML model. Han et al. \cite{han2023reinforcement} show how a model inversion attack against a black-box generative model can be constructed using its confidence scores as rewards to a Reinforcement Learning (RL) agent.

Model extraction is the process of creating a new model that approximates the behavior of the target model. Given enough queries, an adversary can careful craft training data to a surrogate ML model. The effect of this strategy can be amplified if the adversary how some knowledge (gray-box) of the foundational model if the target model has been fine-tuned \cite{li2024shake}. Foundation models enable transfer learning  \cite{bommasani2021opportunities} because they allow their weights to be adapted to new tasks via fine-tuning. The added benefit of using a pre-trained model also comes with risk because they can retain information about the original dataset \cite{lee2024balancing}.

Exfiltation is not restricted to just ML models and can also be used to approximate models tied to first-principles (e.g., physics-based). Similar to targeting ML models, a threat actor with API access to a digital twin that uses a first-principles model could launch a query-based attack to compile enough input-output training pairs to derive a data-driven surrogate model of the target. The use of ML to augment physics modeling is not a novel idea \cite{willard2020integrating, zubatiuk2021development
}. To the best of our knowledge, this is the first research paper to discuss model extraction privacy attack of a physics-based model using a ML surrogate model.

\begin{table*}[ht]
{\renewcommand{\arraystretch}{1.50}% 1.2
%\begin{tabularx}{\textwidth}{X|X|X|X|X|X|X}
%\begin{tabularx}{\textwidth}{X|X|X|X|X|X}
\caption{Privacy attacks and exfiltration techniques against DT-integrated robotic systems adapted from MITRE ATLAS model. \\ (*) Denote proposed additions to the MITRE ATLAS model.}
\vspace{-2mm}
\scriptsize
\begin{tabular}{p{2cm}|p{3.0cm}|p{9cm}|p{1.5cm}} % p{2cm}|p{2.0cm}|p{2.9cm}
%\begin{tabular}{p{3.4cm}|l|l}
\hline
\rowcolor{cyan!20!}
\textbf{Technique} & \textbf{DT Enabling Technology} & \textbf{Overview} & \textbf{Research}\\         
\hline

\multirow{4}{*} {}Exfiltration via Model Inference API &  First-principles* (Physical, Mathematical, Statistical) or ML-enabled models & ML models are prone to leak sensitive information about its training data and adversaries can exfiltrate information through the model API. This can lead to private data leak or model extraction itself~\cite{mitreMlAPIInference}. Non-ML models* are also vulnerable to exfiltration through surrogate creation using ML approaches.  & \cite{alcaraz2022digital, holmes2021digital, scheibmeir2019api, hu2022membership, shokri2017membership, zhu2022label, han2023reinforcement, li2024shake} \\ \hline

\begin{comment}
\cite{alcaraz2022digital, holmes2021digital, minerva2020digital, scheibmeir2019api}

{alcaraz2022digital, holmes2021digital, minerva2020digital, scheibmeir2019api, hu2022membership, shokri2017membership, zhu2022label, han2023reinforcement, li2024shake}
\end{comment}

Exfiltration via Cyber Means & First-principles* (Physical, Mathematical, Statistical) or ML-enabled models & ML artifacts or other information, including non-ML models*, that may be relevant to adversaries' goals could be exfiltrated through traditional cyber means~\cite{mitreMlCyberInference}.  & \cite{alcaraz2022digital, holmes2021digital, rubio2019tracking, saad2020implementation, eckhart2018towards} \\ \hline

LLM Meta Prompt Extraction & LLM-enabled models & An adversary may induce an LLM to disclose its internal instruction prompts, also called ``meta prompt." The leak of meta prompt can equip an adversary with the knowledge of the system's internal workings, and security policies or even may lead to intellectual property theft ~\cite{mitreLlmpromptExtraction}. & \cite{greshake2023not, sha2024prompt, yang2024prsa, wu2024new, wei2024jailbroken, chu2024comprehensive} \\ \hline

\begin{comment}
\cite{greshake2023not, sha2024prompt, yang2024prsa} 
\end{comment}

LLM Data Leakage & LLM-enabled models & An adversary may craft prompts purposely to induce an LLM to disclose sensitive information. Sensitive information can include private user data or proprietary information and may come from proprietary training data, retrieval sources the LLM is connected to, or information from other users of the LLM ~\cite{mitreLlmDataLeak}. & \cite{greshake2023not, liu2023prompt, namer2023cost, evertz2024whispers, kumarstrengthening, carlini2021extracting, huang2022large, birch2023model} \\ \hline

% \begin{comment}
% \cite{greshake2023not, liu2023prompt, namer2023cost, evertz2024whispers, kumarstrengthening, panda2024teach}

%{greshake2023not, liu2023prompt, namer2023cost, evertz2024whispers, kumarstrengthening, panda2024teach, carlini2021extracting, huang2022large, birch2023model}
% \end{comment}
                                    
\end{tabular}}

\vspace{-4mm} %

\end{table*}
\label{table:dt_attacks}

\begin{comment}

%%%% Poisioning %%%%%%
\multirow{3}{*}{Poisoning} & Pre & Attack at model training. Corrupt the ML-based DT by influencing the training data or model parameters \cite{jagielski2018manipulating}, \cite{Kravchik_ics_poisoning_2021}. & Most Physics-based models and simulations will contain adjustable parameters \cite{hasse2017boon}. Corrupt the Physics-based DT by influencing the adjustable parameters. \\ \hline
%%%% end Poision %%%%%%                                    

%%%% Evasion %%%%%%    

Evasion & Post & Attack at model inference. Craft adversarial examples to produce misleading results. \cite{pitropakis2019taxonomy}. &  Use query-based attacks to gather information about the Physics-based DT and craft adversarial examples.\\ \hline

%%%% end Evasion %%%%%%   
\end{comment}

\subsection{Exfiltration via Cyber Means} A DT and its corresponding physical system can also be targeted by traditional cyberattacks. Once adversaries have performed reconnaissance to identify security gaps of a system, they can infiltrate the Robot-DT ecosystem and establish a footprint (e.g., command and control) either through network or physical access (e.g., insider threat). 
{Adversaries may also attempt to exfiltrate data via a physical medium, as illustrated in Fig} \ref{attacks}, {such as a removable drive} \cite{atlasadversarial} {(e.g. external hard drive, USB drive, or other removable storage and processing device)}. Sensitive training data and models can exfiltrated through communication space channels, sometimes in chunks to avoid triggering network traffic thresholds and security mechanisms. This is a well studied research area and readers are directed to the MITRE ATT\&CK framework \cite{strom2018mitre} and recent papers for more insight \cite{lanotte2020formal, alguliyev2018cyber, neupane2023impacts}.

An important addition that this paper provides to the literature is that models based on first principles (i.e., non-ML models) are also susceptible to this attack strategy. Borky et al. \cite{borky2019protecting} advice on the use of cybersecurity practices within a MBSE process in order to protect sensitive information. MBSE and physics-based models tie-back to requirements \cite{bruggeman2022mbse} and known behavior \cite{glatt2021modeling} for the system. Depending on the motivation of the adversary, successful exfiltration of sensitive information allows for extortion for an immediate gain. It also makes the system vulnerable to future attacks for the rest of its lifecycle if the initial attack goes unnoticed and the discovered vulnerabilities go unmitigated.

\subsection{LLM Meta Prompt Extraction} DT-integrated AI robots can leverage Large Language Models (LLMs) to augment the Human-Robot Interaction (HRI) \cite{zhang2023large}. LLMs are well known for their few-shot learning \cite{brown2020language} and In-Context Learning (ICL) capabilities which makes them widely popular and readily accessible to non AI/ML experts. Adept adversaries can attempt to circumvent the safety constraints and safeguarding mechanisms \cite{wu2024new} of a targeted LLM to retrieve its initial instructions (i.e., meta prompt). This technique is sometimes called a \enquote{jailbreaking} attack \cite{wei2024jailbroken} because it allows the adversary bypass guardrails to access internal workings of the model through strategic inputs such as role-playing prompts \cite{chu2024comprehensive}. With the knowledge gained through LLM meta prompt extraction, an adversary can launch additional attacks including theft of sensitive information. If the LLM is a crucial component to the operation and decision-making process of an AI robot, then a successful jailbreaking attack could lead to a compromised system.

\subsection{LLM Data Leakage}
% \ivan{Expand on this section some. Done!}
As described earlier, LLMs are vulnerable to leaking sensitive data \cite{evertz2024whispers} if special care is not taken by integrators. Studies have shown that LLMs are capable of leaking their training data \cite{carlini2021extracting} due to memorization \cite{huang2022large}. Besides training data, LLMs are also capable of leaking connected data sources, information from other users, and model information. Similar to black-box extraction attacks to other ML models, 

Birch et al. \cite{birch2023model} show a \enquote{model leeching} attack that is cost-effective and capable of distilling task-specific knowledge from a target LLM to generate a reduced parameter model. Model leeching is described as a black-box adversarial attack that aims to extract the target LLM by creating a copy (i.e., recreating the target model) within a specific task. To recreate the model, three stages are needed: prompt design, data generation, and  (stolen) model training. If successful model recreation occurs, then adversaries can use it as a surrogate in a staging ground to launch additional attacks against the target model.

Finlayson et al. \cite{finlayson2024logits} show that it is possible to learn a significant amount of non-public information from API-protected LLMs with a small amount of queries. The approach exploits the low-rank output layer common to most LLM architectures. In their research, it shown that the softmax bottleneck imposes low-rank constraints on LLM outputs. They leverage the restricted output space to obtain the LLM image using a small number of LLM outputs. The LLM image can be thought of as a model signature and it exposes model hyperparameters, output layer parameters, and full model outputs.
 
\section{DT-Integrated Robotics Design Considerations and Discussion}
\label{discussion}

In previous sections, we discussed how AI and DT technologies are essential for enabling robots to perform complex tasks in unpredictable environments including safety-critical applications where human-life could be at risk. These technologies can be used in any and all stages of the the AI robot's lifecycle from design (beginning of life) to retirement (end of life) \cite{yousefnezhad2020security} . The use of these technologies also provides a pathway for threat actors to carry out cyberattacks including those that aim to extract private information about the system.

In this section, we touch on the impact of ML model training, responsible AI/DT safeguards, and ethical considerations to the effectiveness of privacy attacks.

\subsection{Risk Factors in ML-enabled Digital Twins}

If a ML-enabled digital twin is used during any phase of an AI robot's lifecycle, special care must be taken in securing those models because factors such as overfitting, dataset structure, model architecture, and model type could affect the accuracy of certain privacy attacks \cite{liu2021machine, rigaki2023survey}.

\subsubsection{ML Model Overfitting and Data Structure}
Overfitting occurs when the underlying ML model in a data-driven DT believes the noise and outliers of the training data to the extent that it performs poorly on new, unseen data. By memorizing details of the training instances, these ML models may not learn the underlying trends in the data and could leave them at risk for privacy attacks. In their research, Yeom et al. \cite{yeom2018privacy} introduce a series of formal definitions for examining the effects of overfitting to membership inference and attribute inference attacks. The theoretical and experimental results of their study show that ML models are more vulnerable to attacks as more overfitting occurs. As part of their study, they analyze linear regression, tree, and Convolutional Neural Network (CNN) models on different datasets (e.g., IWPC, Netflix). Yeom et al. \cite{yeom2020overfitting} later show that overfitting is not a necessary condition to privacy attacks and that models trained to be robust against adversarial examples are also exposed. They note that defending against both privacy and integrity attacks simultaneously may be challenging in some cases.

The composition of the training data can contribute the vulnerability of the ML-enabled DT to privacy attacks. Parallel to overfitting is the notion that if a model has to memorize out-of-distribution samples (i.e., outliers) found in the training data, then those samples are vulnerable to attacks.  In addition, the use Personal Identifiers (PIDs) as part of the model training or inference only increases the privacy attack surface \cite{podoliaka2022privacy}.

\subsubsection{ML Model Complexity}

Similar to overfitting, model complexity can also affect data privacy. Highly complex models, especially deep learning models, contain large number of parameters that could lead them to memorizing the training data instead of learning patterns for generalization. Another added effect is that complex models are usually obscure in terms of interpretability which make finding potential data leaks difficult.

In order to address transparency and to provide the user a level of confidence, AI practioners may augment their models with explainable AI (XAI) techniques. Special care must be taken to balance both explanability and privacy because the use XAI techniques can lead to privacy risks \cite{liu2024please}. With access to explanations, adversaries could craft stronger privacy attacks including member inference and model inversion. Shokri et al. \cite{shokri2021privacy} show that backprogation-based techniques are capable of leaking membership information because of their ability to statistically characterize decision boundaries. Zhao et al. \cite{zhao2021exploiting} develop an image-based, XAI-awared inversion model with emotion prediction as the target task and face reconstruction as the attack task.

The bulk of the privacy attack research is focused on  generative models \cite{chen2020gan, hilprecht2019monte} including Variational Autoencoders and Generative Adversarial Networks. Previously, we discussed how LLMs are also vulnerable to leaking sensitive information. Today, the largest and most capable LLMs are using a transformer-based architecture \cite{yang2024harnessing}. Due to their popularity, there's a rising trend to use transformers for other tasks. Lu et al. \cite{lu2022april} introduce the Attention Privacy Leakage (APRIL) attack to steal private local training data from shared gradients of a Vision Transformer (ViT). The attack showcases the vulnerability of learnable position embeddings.

\subsection{Preventing Privacy Attacks}
While the aim of this paper is to provide an overview of privacy attacks against DT-integrated AI robots, it is also important to discuss defensive strategies against these attacks. There are several defenses against privacy attacks {(or Privacy Enchancing Techniques (PET))} including anonymization, encryption, and differential privacy \cite{vakanski_slides}. The underlying principle behind these defenses {are} tied to \textit{The Fundamental Law of Information Recovery}\cite{dwork2014algorithmic} which formulates the notion that \enquote{giving overly accurate answers to too many questions will inevitably destroy privacy.} The law was formalized and proven with \textit{reconstruction attacks} \cite{dwork2017exposed} which describes any method that aggregates (i.e, compiles) publicly-available information to partially recreate a private dataset. Next, we will describe several major defensive strategies for the privacy attack techniques described in Table \ref{table:dt_attacks}.

For query-based privacy attacks, one simple mitigating approach is to restrict the number of model queries. This prevents adversaries from compiling enough question-answer (QA) pairs to carry out their attack. \textit{Anonymization} techniques can also be employed to remove personable identifiers and information \cite{vakanski_slides} from the training data. This strategy is usually not useful on its own since there are  de-anomynization approaches available to adversaries \cite{majeed2020anonymization}.

\textit{Homomorphic encryption} \cite{li2020privacy} is another defensive strategy against this type of attacks since it allows for computations to be performed on encrypted data. With Homomorphic Encryption (HE), only the user with the matching private key is able to decrypt data to reveal its contents. A practical HE approach has been shown with transformers but it comes with a performance burden \cite{chen2022x}.

\textit{Differential privacy} is closely tied to reconstruction attacks and \textit{The Fundamental Law of Information Recovery}\cite{dwork2014algorithmic}. Differential privacy (DP) can be achieved by adding noise to the training data, model parameters, or model outputs \cite{mireshghallah2020privacy}. Recently, DP has been demonstrated on LLMs \cite{singh2024whispered}.

\textit{Model watermarking} \cite{boenisch2021systematic} could also be used in order protect a model's Intellectual Property (IP) rights. This approach embeds unique identifiers into ML models allowing for traceability back to the model if it is ever falls victim to an exfiltration attack. The watermarking strategy would have to be robust to \enquote{watermark overwriting} attacks that adversaries could employ \cite{zhang2021deep_watermarking}.

In order to secure DTs used in robotics, it is crucial that cybersecurity best practices are used at the onset of their lifecycle. This means that the information and artifacts used from design to deployment are protected using strict access controls and authentication mechanisms. 
\begin{comment}
This includes providing access to such information to only authorized users if the DT is linked to sensitive data at any stage of its lifecycle that could comprise the physical system, stakeholders, or human life. The infrastructure surrounding the DT must contain monitoring and auditing systems to detect suspicious (i.e., anomalous) behavior and to safeguard the DT. As stated before, a comprised DT propagates the attack vector to the physical system, stakeholders, and human life (in safety-critical applications).
\end{comment}
\subsection{Trustworthy, Ethical, and Responsible DT}
\begin{comment}
\ivan{we need probably need to add one subsection here tied to "data governance" which is what Trisha was trying to tackle}
\subash{new addition}
\end{comment}
Data-driven DTs generate predictions, which are then analyzed to extract insights. Stakeholders subsequently use these insights to make decisions, such as whether to perform maintenance or optimize a component. However, the AI models utilized for prediction are typically black-box models, lacking clear explainability. In other words, they are unable to justify their judgments and predictions \cite{9927396}. A trustworthy DT should enable users to comprehend the decision-making process and the rationale behind its actions. One way to enhance trustworthiness in DTs that leverage AI models is to ensure the explainability of these models. This can be achieved through two approaches: employing intrinsically transparent models for prediction, such as decision trees or linear models, or generating explanations \cite{neupane2023twinexplainer} after a decision has been made, also known as post-hoc explanations \cite{kobayashi2024explainable}.

Apart from predictions, AI robots must be trusted to behave safely and ethically. For instance, DTs of autonomous vehicle's (AV) perception systems must have sufficient situational awareness, for it to make the right decisions, to keep itself and humans safe \cite{kuipers2020perspectives}. Isaac Asimov’s three laws of robotics \cite{asimov1941three} can be used as guidelines for programming AI agents within robotic systems to augment trust.

\begin{comment}
Drawing inspiration from Isaac Asimov’s first law of robotics \cite{asimov1941three}: "\emph{A robot may not harm a human being, or, through inaction, allow a human being to come to harm}." Asimov's other two laws are "\emph{A robot must obey the orders given to it by human beings, except where such orders would
conflict with the first law}" and "\emph{A robot must protect its own existence, as long as such protection does not conflict with the first or second law}." These laws can be used a guidelines for programming AI agents within robotic systems to augment trust.
\end{comment}

Ethics has various definitions and approaches in the literature. Kuipers \cite{kuipers2020perspectives} defines ethics as a \enquote{\emph{set of beliefs that a society conveys to its individual members, to encourage them to engage in positive-sum interactions and to avoid negative-sum interactions.}} Robotic systems and applications are governed by \emph{roboethics} \cite{veruggio2005birth}, which oversees the ethical outcomes and aftermaths arising from robotics technology. When it comes to artificial intelligence, AI ethics outlines the moral responsibilities and duties of both the AI system and its developers. The combination of roboethics and AI ethics plays a pivotal role in shaping the development and deployment of responsible DT-Integrated AI robots. Responsible DT can be achieved by considering three levels of AI ethics when designing intelligent systems: \emph{ethics in design, ethics by design,} and \emph{ethics for design} \cite{dignum2019responsible}. The first approach ensures that the development process takes into account the ethical and societal implications of AI and its role in the socio-technical environment. The second approach is concerned with integrating ethical reasoning abilities as part of the behavior of artificial autonomous systems. The third approach, on the other hand, focuses on the research integrity of stakeholders (researchers, developers) and institutions to ensure regulation and certification. Overall, a responsible DT is ethical, lawful, reliable, and beneficial. 

% \hl{subash fixing---with example.----WIP}

% as guidelines for programming AI agents within robotic systems, thereby enhancing their trustworthiness.  The three laws are:

% \begin{enumerate}
% \item A robot may not harm a human being, or, through inaction, allow a human being to come to harm.
% \item A robot must obey the orders given to it by human beings, except where such orders would
% conflict with the First Law.
% \item A robot must protect its own existence, as long as such protection does not conflict with the First
% or Second Law.
% \end{enumerate}
% \ivan{listed the laws above. Remove or keep... thougths?}
\begin{comment}
Additionally, institutions such as The Alan Turing Institute and the University of York are taking initiative in designing a platform known as Trustworthy and Ethical Assurance (TEA) \cite{tea} to foster trustworthiness and ethical assurance for DTs.
\end{comment}

% Furthermore, several initiatives are 

% Trustworthy and ethical assurance is a methodology and procedure for developing a structured argument, which provides reviewable (and contestable) assurance that a set of claims about a normative goal of a data-driven technology are warranted given the available evidence.

% li2023trustworthy \cite{li2023trustworthy}
% hansen2023framework \cite{hansen2023framework}
% jeon2024framework \cite{jeon2024framework}

% \ivan{Isaac Asimov's laws, "Three Laws of Robotics"}

\subsection{Data Governance and Data Management}

AI-enabled digital twins, including those used in robotics, must be managed and protected using \textit{data governance} practices and processes. In fact, data governance (DG) lays the foundation for Trustworthy AI \cite{janssen2020data}. The purpose of DG is to build trust in data \cite{eryurek2021data}. It encompasses the policies, standards, and practices required to effectively manage data throughout its lifecycle. The use of data governance frameworks ensures that data is accurate, available, and secure.

Transparency is crucial in order to build trust and accountability with stakeholders. Important questions that need to be addressed in data-driven processes include: \emph{What are the types of data being collected? How will the data be used? Who will have access to the data?} Organizations involved in DT typically establish contractual agreements and develop data governance frameworks to clarify \textit{data ownership}. Compliance with data protection regulations, such as GDPR \cite{li2019impact_gdpr} or the CCPA \cite{goldman2020introduction},  may require data ownership declarations and obtaining consent for data collection/usage in order to ensure transparency.

Tied to transparency is the concept of traceability \cite{mora2021traceability}. The data collected by DTs, processed by (e.g., features) DTs, and made by DTs (e.g., decisions, inferences) must be logged to allow for traceability. The data transformations that occur in AI robots throughout data-driven processes are operating on behalf of individuals or making autonomous decisions. For these decisions to be ethically and responsibly taken, they must be explainable and align with transparency requirements which ensures fairness and democracy ~\cite{holler2016digital,white2021digital,Chang2022transparent}. This is especially true in safety-critical applications where loss of life or the environment may be damaged if the AI robot fails. \emph{Who is responsible when AI robots fail? How can we accurately determine what went wrong when the system fails?}

\subsection{Trusted Autonomy}
\label{trusted_autonomy}
Trusted autonomy refers to the design, development, and deployment of AI robots that can be relied upon to operate safely, effectively, and ethically within their intended environments. Experts \cite{trusted_autonomy} describe research in trusted autonomy as covering multiple \enquote{advanced topics such as robotics, AI, simulation and ethics.} Despite significant progress, achieving trusted autonomy is an ongoing effort that requires continuous improvement and adaptation to new challenges.

We believe that trusted autonomy can be achieved by combining robotics with AI and DT technologies. In addition, ethics and trustworthiness procedures and cybersecurity processes are needed. AI allows for efficiently searching high-dimensional spaces \cite{perez2020membrane} and pattern recognition. The latter allows for Dimensionality Reduction (DR) \cite{jia2022feature} and data compression \cite{liu2021high}. It is hard for humans to visualize data beyond a low number of dimensions (e.g, 3D). Existing techniques along with the use topological AI and Topological Data Analysis (TDA) \cite{chazal2021introduction} can help reduce a high dimensional problem space. In doing so, it improves the explainability of these data-driven approaches allowing stakeholders to better visualize why decisions are being made.

AI and digital twin models can be integrated with Modeling and Simulation (M\&S) frameworks allowing for \enquote{What-If} scenarios. Stakeholders can use make use of a digital twin in lieu of the physical system to explore and analyze how the asset would act in simulated environmental configurations. In other words, the use of a DT allows for Testing and Evaluation (T\&E) of the physical system if the models provide the needed fidelity. This is extremely important in safety-critical applications where human-life, the environment, or the system must be protected.

Data governance is the connective tissue that ties accuracy, availability, and security of data in the data-driven models and processes used by AI robots. Cybersecurity principles must also be used to ensure that models are secure and protected from adversaries looking to do harm (e.g., obtain unauthorized access to sensitive information). The merger of all the previous discussed concepts leads us into responsible and ethical use of AI robots and closer to trusted autonomy.

\section{Conclusion}
\label{conclusion}
{The convergence of AI and DT technologies presents both opportunities and challenges for the development of complex robots. As these robots become increasingly integrated into safety-critical applications, ensuring their security and trustworthiness is paramount. As illustrated in our survey, both ML-enabled and physics-based robotic DT models are susceptible to various privacy attacks aimed at extracting sensitive information, potentially compromising not only intellectual property but also human safety. To address these challenges, we advocate for a holistic approach that integrates cybersecurity best practices, robust privacy-preserving techniques, and adherence to ethical principles throughout the robot's DT lifecycle. Furthermore, we highlight the importance of the trusted autonomy,  where the design and operation of these robots prioritize transparency, explainability, and rigorous security measures, ensuring they operate safely, responsibly, and transparently.  The development and deployment of trustworthy, ethical, and secure DT-integrated AI robots will be crucial for their successful adoption in various sectors.}

\section*{Acknowledgements}

This work was supported by the Predictive
Analytics and Technology Integration (PATENT) Laboratory at the
Department of Computer Science and Engineering, Mississippi State University. The authors would like to acknowledge Dr. Jorge A. O'Farrill and Dr. Stephen R. Snarski (Technical Fellows, Modern Technology Solutions, Inc.) for their contributions to both Digital Twin and Trusted Autonomy topics.

\bibliographystyle{IEEEtran}
\bibliography{IEEEabrv, refs}

\begin{comment}
\bibliographystyle{unsrt}
\bibliography{refs}
\end{comment}

\end{document}